\documentclass[letterpaper, 10 pt, conference, nofonttune]{ieeeconf} % Comment this line out if you need a4paper

\IEEEoverridecommandlockouts                              % This command is only needed if
\usepackage{amsthm}

\usepackage{subfigure}
\usepackage{xcolor}
\usepackage{times}
\usepackage{epsfig}
\usepackage{amsmath}
\usepackage{amssymb}
\usepackage{mathrsfs}
\usepackage[ruled,vlined,linesnumbered]{algorithm2e}
\usepackage{wrapfig}
\usepackage{hyperref}

\newcommand{\GGG}{\mathcal{G}}

\newcommand{\VVV}{\mathcal{V}}

\newcommand{\RRR}{\mathcal{R}}
\newcommand{\NNN}{\mathcal{N}}

\newcommand{\GG}{\mathbf{G}}
\newcommand{\SIM}{\mathbf{S}}
\newcommand{\HHH}{\mathbf{H}}
\newcommand{\hh}{\mathbf{h}}

\newcommand{\vv}{\mathbf{v}}
\newcommand{\qq}{\mathbf{q}}
\newcommand{\kk}{\mathbf{k}}

\newcommand{\pp}{\mathbf{p}}
\newcommand{\AAAA}{\mathbf{A}}
\newcommand{\DD}{\mathbf{D}}
\newcommand{\XX}{\mathbf{X}}
\newcommand{\one}{\mathbf{1}}

\newcommand{\MM}{\mathbf{M}}

\newcommand{\WW}{\mathbf{W}}

\newcommand{\UU}{\mathbf{U}}

\newcommand{\YY}{\mathbf{Y}}
\newcommand{\RR}{\mathbb{R}}
\newcommand{\xx}{\mathbf{x}}

\makeatother

\title{\LARGE \bf
Deep Masked Graph Matching for\\ Correspondence Identification in Collaborative Perception
}

\begin{document}
\author{Peng Gao$^{1*}$, Qingzhao Zhu$^{2*}$, Hongsheng Lu$^{3}$, Chuang Gan$^{4}$, and Hao Zhang$^{5}$
% \thanks{$^*$This work was partially supported by ....----------------------------------------------------}
\thanks{Authors with $*$ contributed equally to this paper.}
\thanks{$^{1}$Peng Gao is with the Department of Computer Science, University of Maryland, College Park, MD 20742, USA. {Email: gaopeng@umd.edu}.}%
\thanks{$^{2}$Qingzhao Zhu is with the Department of Computer Science, Colorado School of Mines, Golden, CO 80401, USA. Email: zhuqingzhao@mines.edu.}%
\thanks{$^{3}$Hongsheng Lu are with Toyota Motor North America, Mountain View, CA 94043, USA. Email: hongsheng.lu@toyota.com.}%
\thanks{$^{4}$Chuang Gan is with MIT-IBM Watson AI Lab, Cambridge, MA 02142, USA. Email: ganchuang@csail.mit.edu.}%
\thanks{$^{5}$Hao Zhang is with the Manning College of Information and Computer Sciences (CICS),
University of Massachusetts Amherst, Amherst, MA 01002, USA.
        {Email: hao.zhang@cs.umass.edu}.}
}

\maketitle
\thispagestyle{empty}
\pagestyle{empty}

\begin{abstract}
Correspondence identification (CoID) is an essential component for collaborative perception in multi-robot systems, such as connected autonomous vehicles.
The goal of CoID is to identify the correspondence of objects observed by multiple robots in their own field of view in order for robots to consistently refer to the same objects.
CoID is challenging due to perceptual aliasing, object non-covisibility, and noisy sensing.
In this paper, we introduce a novel deep masked graph matching approach to
enable CoID and address the challenges.
Our approach formulates CoID as a graph matching problem
and we design a masked neural network to integrate the multimodal visual, spatial, and GPS information to perform CoID.
In addition, we design a new technique to explicitly address object non-covisibility caused by occlusion and the vehicle's limited field of view.
We evaluate our approach in a variety of street environments using a high-fidelity simulation that integrates the CARLA and SUMO simulators.
The experimental results show that our approach outperforms the previous approaches
and achieves state-of-the-art CoID performance in connected autonomous driving applications. Our work is available at: \url{https://github.com/gaopeng5/DMGM.git}.
\end{abstract}

\section{Introduction}

Multi-robot systems have been widely investigated over the past decades due to their reliability and efficiency to address collaborative tasks,
such as collaborative manufacturing \cite{gao2021bayesian, zhang2020real},
multi-robot-assisted search and rescue \cite{reily2021balancing, robin2016multi}, 
and connected autonomous driving \cite{guo2019collaborative, wei2018survey}. 
To enable effective multi-robot collaboration, 
collaborative perception is a fundamental capability for multiple robots to share their perceptual data of the surrounding environment 
and to build a shared situational awareness among the robots.

\begin{figure}
\centering
\vspace{6pt}
\includegraphics[width=0.485\textwidth]{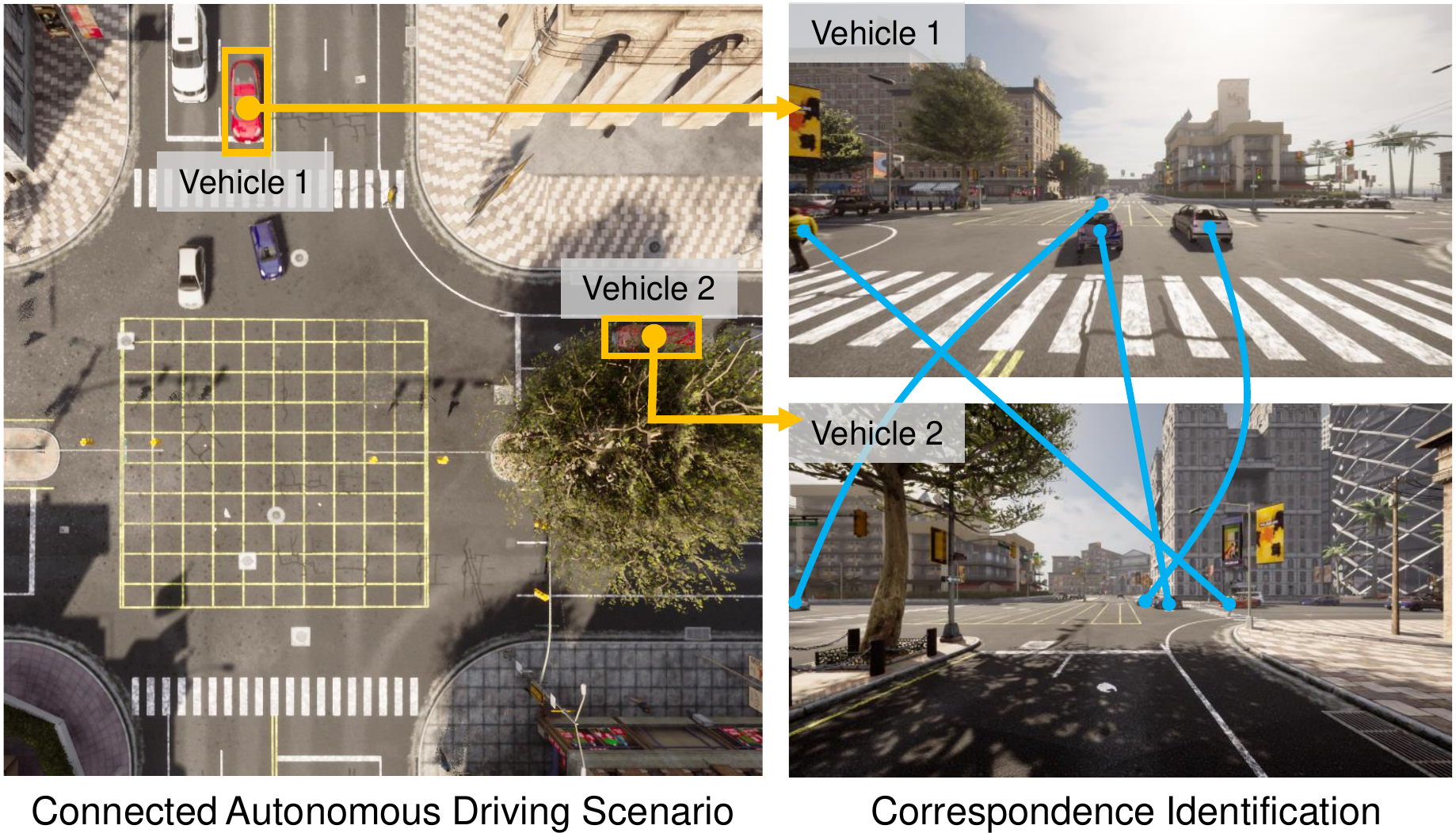}
%\vspace{-6pt}
\caption{A motivating scenario of correspondence identification to enable collaborative perception in connected autonomous driving.
When two connected vehicles meet at an intersection,
before they can share information of street objects,
they need to identify the correspondence of the objects observed in their own field of view
in order to consistently refer to the same objects.
}\label{fig:motivation}
\vspace{-6pt}
\end{figure}

As a critical component of collaborative perception, 
correspondence identification (CoID) aims to 
find the correspondence of the same objects observed by multiple robots in their own field of view. 
Figure \ref{fig:motivation} demonstrates an example of CoID.
When a pair of connected vehicles meet at an intersection, 
they have to identify the correspondence of the street objects in order to correctly refer to the same objects when they share information about the objects.

Given the importance of CoID, 
a variety of approaches are developed, including learning-free and learning-based methods. 
Learning-free methods can be divided into three groups, 
including keypoint-based visual association \cite{arandjelovic2016netvlad}, 
geometric-based spatial matching \cite{gao2021regularized,gao2020correspondence}, 
and synchronization methods that perform multi-view data association \cite{fathian2020clear, boyd2011distributed}. 
Learning-based approaches are mainly based on deep learning, 
such as convolution neural network to perform object re-identification from different perspectives \cite{yu2018unsupervised, sarlin2020superglue}
and graph neural network to perform deep graph matching \cite{wang2019learning,gao2022correspondence}.

CoID is a challenging problem due to several reasons. 
First,
enabling CoID must address the challenge of perceptual aliasing, 
i.e., street objects having similar or identical appearances, which often introduce visual ambiguity.
The second challenge is caused by non-covisible objects that are only observed by a single robot 
due to occlusion or the robot's limited field of view. 
% which will lead to significant number of mismatches (true negative). 
The third challenge is caused by noisy perception.
For example, noisy depth sensing often causes inaccurate distance estimation.

To address the above challenges, we propose a novel deep masked graph matching method to perform CoID.
We develop graph representations to encode visual-spatial information of street objects observed by each vehicle.
Each node in a graph represents a street object and each edge represents the spatial relationship of a pair of objects.
Given the graph representations of observations obtained by a pair of connected vehicles, 
we mathematically formulate CoID as a graph matching problem,
and we develop a new masked graph neural network that integrates visual, spatial, and GPS cues
to explicitly address object non-covisibility.

% Finally, we create a simulator based on CARLA, which can simulate
% connected autonomous driving and provide large amount of training data.

The key contribution of this paper is the introduction of the masked deep graph matching method for CoID in collaborative perception. 
% that integrates multi-modal information with novel loss and non-covisible object remover for correspondence identification. 
Specific novelties include
%(1) a multi-modal graph-based representation that is able to integrate visual, spatial and GPS information to improve the representativeness 

\begin{itemize}
    \item We propose a multi-modal graph-based formulation that is able to integrate visual, spatial and GPS information
     to better represent  street objects to improve CoID.
     
    \item We introduce a masked deep neural network with a novel loss design and non-covisible remover based on SoftMax variance thresholding to address object non-covisibility and improve CoID precision.

\end{itemize}

% The remaining of this paper is organized as follows. Section \ref{sec:related} reviews related research on collaborative perception in multi-robot systems and CoID methods.
% Our approach is introduced in Section \ref{sec:approach}.
% In Section \ref{sec:experiment}, we show experimental results.
% Finally, the paper is concluded in Section \ref{sec:conclusion}.

\section{Related Work}\label{sec:related}

\subsection{Collaborative Perception}
Given the popularity of multi-robot systems, collaborative perception attracts attention in recent studies.
One of the well-known uses of collaborative perception is inter-robot loop closure detection in collaborative simultaneous localization and mapping (CSLAM), in which multiple robots need to recognize the same place by identifying the correspondences of landmarks or key points in order to merge their local maps \cite{arandjelovic2016netvlad,gao2020long}.
In addition, collaborative object localization can achieve better performance compared with single-view localization, in which multiple robots require to consistently localize the same objects given their own observations by  identifying the correspondences of objects \cite{gao2022asynchronous, ji2017surfacenet, gao2021multi}.
Furthermore, by associating multi-robot observations, collaborative perception also happens in trajectory forecasting \cite{zhu2021learning}, scene segmentation \cite{liu2020who2com, liu2020when2com}, tracking and object detection \cite{robin2016multi}. In these applications, CoID plays an important role in collaborative perception, with the goal of associating key points or objects in various observations provided by multiple robots. However, most of the existing work assumes that the correspondences of objects are known or can be easily obtained via coordination transformation, such as based on GPS \cite{liu2020who2com, liu2020when2com} or point cloud registration \cite{zhang2014loam}.  The real-world cases without accurate GPS information or accurate poses have not been well studied yet to address CoID in collaborative perception.

In addition, there exist several open-source datasets on collaborative perception. 
CoMap uses CARLA \cite{dosovitskiy2017carla} and SUMO \cite{krajzewicz2002sumo} to generate a large-scale dataset for LiDAR-based object detection and semantic segmentation \cite{yuan2021comap}.
Similar datasets include OPV2V \cite{xu2022opv2v} and DAIRV2X \cite{yu2022dair}, which use LiDAR points to evaluate object detection.  
V2X-Sim is a multi-modal multi-task dataset, which is used to evaluate object detection, segmentation and tracking \cite{li2022v2x}.
Although these datasets provide ground truth for various collaborative perception tasks, 
they do not provide correspondences of street objects in multi-perspective observations obtained by connected vehicles.
Our dataset provides the CoID ground truth that is computed from instance-level semantic segmentation in the simulations.

\subsection{Correspondence Identification}

CoID can be generally grouped into two categories, including learning-free and learning-based approaches.

{Learning-free} approaches can further be divided into three subgroups using visual appearance features, spatial relationship, and synchronization algorithms, respectively. Visual appearance features are commonly used for key-point matching to register adjacent frames or local-global mapping, such as SIFT \cite{engel2014lsd} and ORB \cite{mur2015orb}. In addition, region-based visual features, such as HOG \cite{dalal2005histograms} and TransReID \cite{he2021transreid}, are generally used to identify the same place observed at different times. Besides using visual features, spatial features are also used to identify correspondences of objects, such as ICP \cite{rusinkiewicz2001efficient}, graph matching \cite{gao2020correspondence, gao2021regularized} and maximum clique \cite{adamczewski2015discrete}. Furthermore, synchronization algorithms are also highly related to the problem of CoID \cite{fathian2020clear,boyd2011distributed}. These algorithms take the pairwise correspondences as inputs, and output multi-view correspondences by forcing circle consistency, e.g., using graph cut \cite{fathian2020clear} and convex optimization \cite{hu2018distributable}.

{Learning-based} approaches typically focus on using deep neural networks to perform CoID. 
Specifically, these methods are typically based on convolution neural networks (CNN) or graph neural networks (GNN).
CNN-based methods focus on extracting high-level visual features to recognize the same objects observed from different perspectives \cite{jin2020semantics,khatun2020semantic, voigtlaender2019mots}. GNN-based approaches aim to learn a unique pattern surrounding objects by aggregating their spatial relationships \cite{wang2019learning,zhang2019deep, fey2019deep}.
In addition to the pure CNN or GNN-based approaches, there are several methods using the combination of them \cite{gao2021bayesian, gao2022correspondence}. By integrating visual-spatial information for CoID, it can improve robustness.

Although existing methods show promising performance, they often cannot fuse multi-modal information (visual, spatial, and GPS cues) for CoID.
The problem of non-covisibility has also not been well addressed yet to enable CoID in connected driving scenarios.

\section{Approach}\label{sec:approach}

\textbf{Notation.} Matrices are denoted as boldface capital letters, e.g., $\MM \!=\! \{\MM_{i,j}\}\in \RRR^{n\times m}$.
$\MM_{i,j}$ denotes the element in the $i$-th row and the $j$-th column of $\MM$, 
and $\MM_{i,:}$ denotes the $i$-th row of $\MM$.
Vectors are denoted as boldface lowercase letters $\vv \in \RRR^{n}$, and scalars are denoted as lowercase letters.

\subsection{Problem Formulation}

% We introduce a masked graph neural network to address correspondence identification.
Given an observation observed by a vehicle, we represent it as a graph  $\GGG(\VVV, \XX, \AAAA)$.
$\VVV=\{\vv_1,\vv_2,\dots,\vv_n\}$ represents the node set with $\vv_i \in \VVV $ denoting the 3D position of the $i$-th detected object 
(e.g. vehicle or pedestrian).
Each object is associated with a visual feature vector, denoted as $\XX = \{\xx_1,\xx_2,\dots,\xx_n \}$,
where $\xx_i \in \mathbb{R}^{ d}$ denotes the visual feature vector of the $i$-th object,
and $d$ is the vector length.
$\AAAA$ is the adjacent matrix indicating the node connection.
All nodes are connected via Delaunay triangulation.
If node $i$ and node $j$ are connected, then $\AAAA_{i,j}=||\vv_i-\vv_j||^2$; otherwise, $\AAAA_{i,j}=0$.

In collaborative perception, observations observed by a pair of vehicles can be represented as $\GGG$ and $\GGG'$, respectively. 
The correspondences of the objects observed by the two vehicles are represented by the correspondence matrix $\YY \in \RRR^{n\times n'}$. 
Then, we mathematically formulate CoID as a graph matching problem, which is defined as follows:
\begin{equation}\label{eq:prob_form}
    \arg\max_\YY \SIM^\top \YY \quad
    \textrm{s.t.} \;  \YY\one_{n'\times 1} \leq \one_{n\times 1},\YY^\top\one_{n\times 1} \leq \one_{n'\times 1}
\end{equation}
where $\SIM$ encodes the similarity between $\GGG$ and $\GGG'$ and $\one$ is a all-one vector. 
Given Eq. (\ref{eq:prob_form}), $\YY$ is optimal when the similarity $\SIM$ is maximum according to the identified correspondences. 
The constraint is used to force the one-to-one correspondences. 
In other words, an object in one observation can at most have one corresponding object in the other observation. 
The main technical problem we need to address is how to calculate the similarity $\SIM$ between a pair of graphs $\GGG$ and $\GGG'$.
%the correspondence identification problem turns into a graph matching problem.
%The approach is broadly categorized into three parts:
%graph embedding, graph pruning, and graph matching.
%The first part embeds observations through attention-based graph neural network.
%And the node embeddings are compared by similarity matrix.
%The second part prunes the similarity through both graph structure and GPS information.
%The final part optimizes the similarity and derives correspondence between source and target observations.
%The overall structure of our approach is shown in figure \ref{fig:approach}.

\subsection{Attentional Graph Embedding}
To calculate the similarity between a pair of graphs, we do not just encode each object's own visual features but also its surrounding objects' appearance features given their spatial relationships. Formally, we use an attentional graph neural network  to compute node embedding vectors as  $\HHH=\Psi(\XX, \AAAA)$, where $\HHH$ denotes the embedding matrix and $\Psi(\XX, \AAAA)$ denotes the attentional graph neural network. Each row of $\HHH_{i,:}$ represents an embedding vector of the $i$-th object, which is defined as $\hh_i$.

\begin{equation}
    \qq_i^l = \WW_{q}^l \hh_i^l, \quad   \kk_i^l = \WW_{k}^l \hh_i^l, \quad
    \vv_i^l = \WW_{v}^l \hh_i^l 
\end{equation}
where $l=1,2,\dots,L$ denotes the layer index, $\qq_i^l$, $\kk_i^l$ and $\vv_i^l$ denote query, key and value, $\WW_{q}^l, \WW_{k}^l, \WW_{v}^l$ denote their associating trainable weights. $\hh_i^l$ denotes the feature embedding vector of the $i$-th object at the $l$-th layer, where $\hh_i^0=\xx_i^0$. In the self-attention mechanism,
the object feature $\xx_i^l \in \XX$ is first linearly transformed to query, key and value. Then the
self-attention is computed as follows:
\begin{equation}
    \alpha_{i,j}^l = \textrm{SoftMax}\left( \frac{(\qq_i^l)^\top (\kk_j^l + \WW_{e}^l \AAAA_{i,j})}{\sqrt{c^l}} \right)
\end{equation}
where $\alpha_{i,j}^l$ is the attention from node $j$ to node $i$ at layer $l$. To encode spatial relationships of objects, we add edge attributes into the learning process, where
$W_e^l$ denotes the learnable parameter matrix that has
the same dimension of key $k_j^l$.
$c^l$ is the number of output channels.
This attention weight is obtained by comparing the query of the $i$-th node with
its neighborhood keys and edge attributes. The final attention is normalized by the SoftMax function.
The final node embedding vector is computed as follows:
%by combining the ego information and aggregation of neighboring information.
\begin{equation}
\hh_i^{l+1} = \WW_x^l \hh_i^l + \sum_{\AAAA_{i,j}=1}\alpha_{i,j}^l (\vv_j^l + \WW_{e}^l \AAAA_{i,j})
\end{equation}
$\WW_x^l$ is a learnable parameter matrix. The final embedding vector $\hh_i$ is computed via aggregating the object embedding feature and its neighborhood edge attributes weighted by attention weights.
We also use a multi-head mechanism \cite{velikovi2017graph} to enable the network to catch a richer representation of the embedding. Multi-head embedding vectors are concatenated after intermediate attention layers and averaged after the last attention layer.
% Source and target observations are organized in pair graphs $G_s(V_s, X_s, E_s, A_s)$
% and $G_t(V_t, X_t, E_t, A_t)$.
% Nodes in both graphs are embedded through $\Psi$.
% Assume there are $m$ objects detected in source observations
% and $n$ objects detected in target observations.
% The dimension of embedding of each node coming out of $\Psi$ is $h$.
% The outputs are denoted as source embeddings $O_s \in \mathbb{R}^{m \times h}$
% and target embeddings $O_t \in \mathbb{R}^{n \times h}$.
% The similarity $M$ between two embeddings are obtained by dot products.
Given the object embedding vectors, we compute the similarity between two graphs $\GGG$ and $\GGG'$ as follows:
\begin{equation}
    \SIM_{i,j} =\hh_i\hh_j^\top
\end{equation}
where $\SIM \in \mathbb{R}^{m \times n}$ represents the similarity between the two graphs with $m$ and $n$ objects respectively.

\begin{figure}[t]
\centering\vspace{6pt}
\includegraphics[width=0.47 \textwidth]{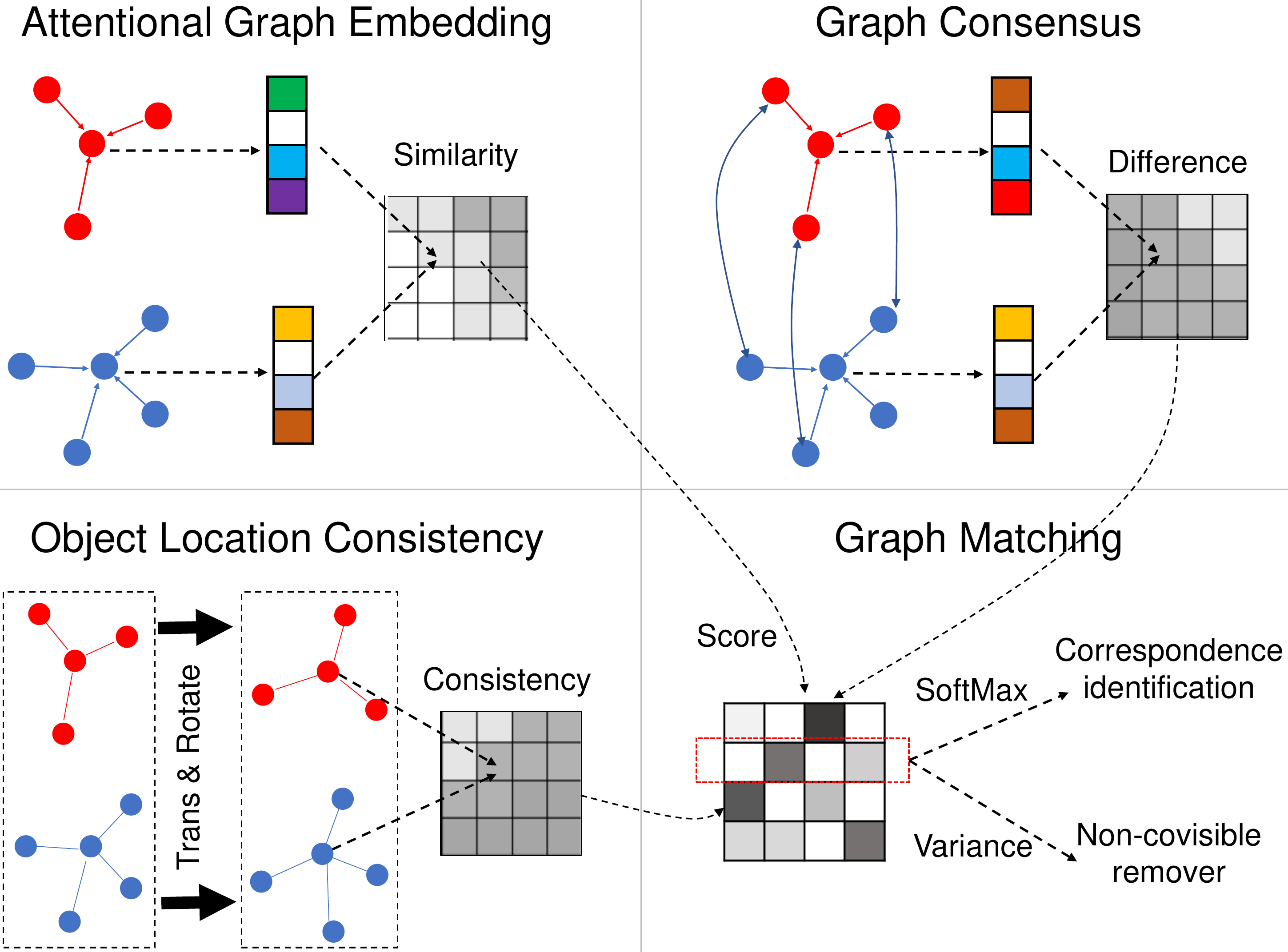}

\caption{Overview of the proposed masked deep graph matching approach for correspondence identification. % 
}
\label{fig:approach}
\vspace{-6pt}
\end{figure}

\subsection{Graph Consensus}
Due to the existence of noise in sensing information, such as noise in depth or RGB observations, it will introduce ambiguity into the CoID process based on visual and spatial information. Thus, we study the graph pruning technique to improve the robustness of CoID from two aspects. Inspired by the most recent work \cite{fey2019deep}, we apply the graph consensus principle to the graphs according to the identified correspondences of objects. Formally,
\begin{equation}\label{eq:update}
    \SIM_{i,j} =\hh_i\hh_j^\top +\varphi(\DD_{i,j})
\end{equation}
where $\varphi$ denotes a multi-layer perceptron with two linear layers followed by a ReLu non-linear activation function. $\DD$ denotes the consensus difference between graphs $\GGG$ and $\GGG'$ given the correspondences $\YY$. Formally,
\begin{equation}
    \DD = (\SIM^\top \Psi(\UU, \AAAA) - \Psi(\SIM^\top \UU, \AAAA'))^\top
\end{equation}
where $\UU \in \RRR^{n \times r}$ is a random matrix with each row $\UU_{i:}$ denoting a random feature vector with length $r$. When $\GGG$ and $\GGG'$ indicate the same graph (isomorphism), then $\SIM^\top \Psi(\UU, \AAAA)= \Psi(\SIM^\top\UU, \SIM^\top \AAAA \SIM)= \Psi(\SIM^\top \UU, \AAAA')$. In this case, $\DD_{i,j}=\mathbf{0}$. Otherwise, $\varphi(\DD)$ that indicates the differences between two graphs will update the similarity matrix $\SIM$.

\begin{figure*}[t]
\vspace{6pt}
\centering
    \includegraphics[width=0.99\textwidth]{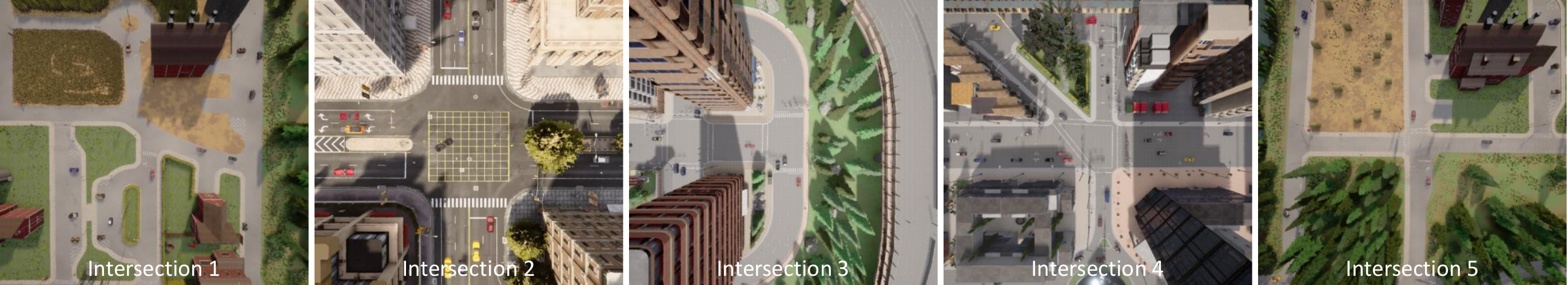}
    \caption{The five intersection scenarios that are implemented in our CAD simulator and used for approach evaluation in the experiments.}
    \label{fig:scenario}
\vspace{-6pt}
\end{figure*}

\subsection{Object Location Consistency}
In addition to pruning graph matching results given the consensus principle, our approach also incorporates GPS information to improve graph matching performance.
We assume that the GPS information of two connected vehicles can be represented as the extrinsic parameters of cameras mounted on the vehicles with respect to the global coordinate. Then we can transfer the objects' 3D positions from the camera coordinates to the GPS global coordinates given the extrinsic parameters.
We represent the positions of objects in the world 3D coordinates as $\{\pp_i\}^n$ and $\{\pp'_j\}^{n'}$ separately.
Given the consistency of object positions in the world coordinates, we construct a position mask $\GG \in \mathbb{R}^{n \times n'}$ based on the distances of corresponding objects in two different graphs.
Each element in $\GG$ can be calculated as:
\begin{equation}\label{eq:gi,j}
    \GG_{i,j} = \frac{1}{\pp_{i} - \pp'_{j}}
\end{equation}
where $\GG$ is calculated via a reciprocal function, which represents the similarity between the positions of the corresponding objects in the world coordinates.
If the corresponding objects denote the same object, then
$\GG_{i,j}$ will be large as their world coordinates are ideally the same.
%In reality, as GPS information is not $100\%$ accurate, and the 3D positions of objects contain various uncertainties, such as sensor resolution, depth estimate error, etc,
The final similarity score $\SIM \in \mathbb{R}^{n \times n'}$ is defined as follows:
\begin{equation}
    \SIM_{i,j} =\hh_i\hh_j^\top +\varphi(\DD_{i,j}) +\GG_{i,j}
\end{equation}
where the similarity score $\SIM$ containing
similarities of visual-spatial features of objects, neighborhood structure consensus, and GPS-based object position consistency, as shown in Figure \ref{fig:approach}. Finally, the correspondences of objects can be identified as follows:
\begin{equation}
    \YY=\text{SoftMax}(\SIM)
\end{equation}
Even though SoftMax can not enforce one-to-one constraints on the correspondences, it has better consistency with the consensus pruning \cite{fey2019deep} and has similar performance as the one-to-one constrained assignment problem solver, such as Sinkhorn normalization.

\subsection{Addressing Non-Covisible Objects}
There may exist many non-covisible objects in two observations. To address this challenge, we propose a threshold-based approach to remove potential non-covisible objects based on SoftMax variance. Specifically, we do thresholding on the standard deviation of each row of $\SIM$, which is defined as $s_i=\sigma(\SIM_{i,:})$, 
%\begin{equation}
%s_i=\sigma(\SIM_{i,:})
%\end{equation}
where $\sigma$ denotes the variance operator and $s_i$ indicates the variance of the $i$-th identified correspondence. If the network is confident in the classification result computed via SoftMax, $s_i$ should be large, otherwise; it should be small (as all the categories have similar confidence in the classification results). Then, we do thresholding on the variance.
If $s_i \ge \theta$, then we preserve the $i$-th object in the correspondence matrix.
If $s_i < \theta$, then we remove it from the identified correspondences by setting $\YY_{i,:}=0$. Given the threshold on the SoftMax variance, we can significantly remove the non-covisible object from the identified correspondences, as they usually have low variance (even confidence) in the classification results.
To train our network, we design the loss function as follows:
\begin{equation}
    \mathcal{L} = \frac{1}{NN'} \sum_{i,j}(\SIM_{i,j} - \YY^*_{i,j})^2
\end{equation}
where $\YY^* \in \mathcal{R} ^{n \times n'}$ denotes the ground truth correspondence. $\YY_{i,j}=1$ denotes that the $i$-th object in one observation corresponds to the $j$-th object in the other observation. Otherwise, $\YY_{i,j}=0$.
The correspondence matrix $\YY \in \mathcal{R} ^{m \times n}$ is optimal when the loss function is minimized.

\section{Experiments}\label{sec:experiment}

In this section, we discuss our experimental setup, results in high-fidelity CAD simulations, and analysis of our approach.

\subsection{Experimental Setup}

We implement a high-fidelity connected autonomous driving (CAD) simulator
that integrates CARLA \cite{Dosovitskiy17} and SUMO \cite{krajzewicz2002sumo}.
CARLA is an open-source autonomous driving simulator that is able to simulate vehicle sensors, driving control and traffic scenarios. 
In our experiments, we design five different traffic scenarios at street interactions where connected vehicles more frequently meet from different driving directions.
These five simulated scenarios are depicted in Figure \ref{fig:scenario}.

\begin{figure}[th]
\centering
    \includegraphics[width=0.485\textwidth]{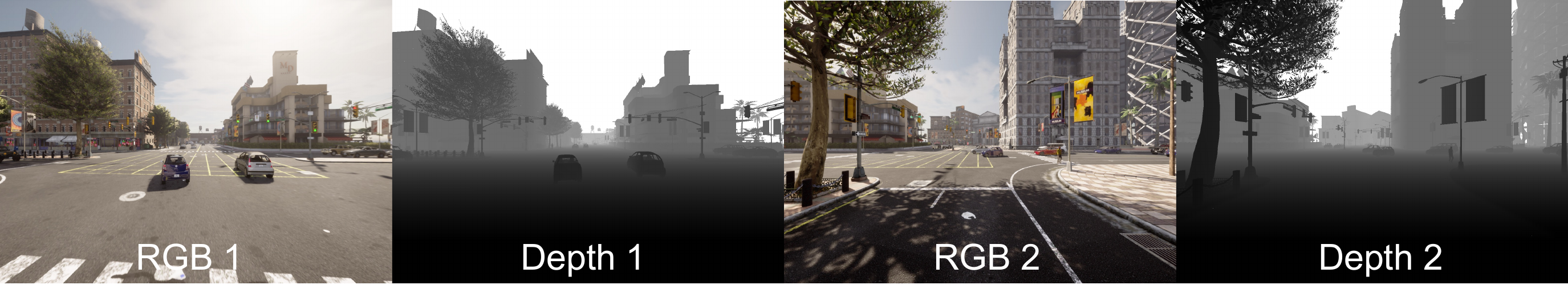}
    \caption{Examples of the simulated color and depth images obtained by two connected vehicles at the same intersection from different perspectives.}
    \label{fig:rgbd}
\end{figure}

In the simulations, each connected vehicle is equipped with a front-facing RGB camera, a front-facing depth camera, and a global navigation satellite system (GNSS) sensor. 
Examples of the RGB and depth images from a pair of connected vehicles are shown in Figure \ref{fig:rgbd}.
Simulation of the GNSS sensor follows the technical specification of the real SBG Ellipse2-D sensor.
Traffic patterns, including pedestrians and vehicles, are controlled by SUMO. 
Behaviors of vehicles and pedestrians are generated randomly and follow real-world rules, such as stopping at the red light and yielding to the pedestrian.

\newcommand{\tabincell}[2]{\begin{tabular}{@{}#1@{}}#2\end{tabular}}
\begin{table}[ht]
    \vspace{4pt}
	\centering
	\caption{Description of our CAD dataset based on CARLA and SUMO simulations for CoID evaluation.}
	\label{tab:dataset}
	\tabcolsep=0.25cm
	\begin{tabular}{|c|c|}
		\hline
		\# Instances & 69,469 from 5 different scenarios \\
  		\hline
		Sensor  & Color, depth, and GNSS sensors  \\
		\hline
%		RGBD FOV & $120^\circ$ \\
%		\hline
		RGBD specs & $1920 \times 1080$ at 10 FPS \\
		\hline
		GNSS noise & \tabincell{c}{$\NNN(0,1.2m)$ in vertical and horizontal directions \\ $\NNN(0,0.2^\circ)$ in yaw}\\
		\hline
	\end{tabular}
\end{table}

Using our CAD simulator, we collect a large CAD dataset that is summarized in Table \ref{tab:dataset}.
We collect a total of 69,469 data instances, of which 60,260 data instances are used for training, 3,747 data instances are used for validation, and 5,462 data instances are used for testing. Each data instance includes a pair of RGBD images observed by two connected vehicles from different perspectives, the GNSS positions and orientations of vehicles, and the ground truth of object correspondences directly obtained from the instance-level segmentation provided by CARLA (each object segmentation has a unique ID).

In our graph construction, we use YOLOv5 \cite{glenn_jocher_2022_6222936} to detect objects and we extract the HOG feature \cite{dalal2005histograms} as each node's appearance feature. The edges are connected via Delaunay triangulation. The edge attributes (distances) are calculated from pairs of objects' positions, which are obtained from the depth images. The GNSS  positions and orientations of each vehicle are represented as the XYZ-pitch-roll-yaw form, which can be rewritten as the transformation matrix.

In the implementation of our network, the attentional GNN $\Psi$ is implemented based on the PyTorch geometric library. 
We set the number of network layers to be $L=2$. In the first network layer, we set  $\WW_q,\WW_k,\WW_v \in \RR ^{900\times \{heads * 256\}}$ where the multi-head number $heads=4$. In addition, we set $\WW_e \in \RR^{ dim \times \{heads * 256\} }$ where the edge feature dimension $dim=1$. Each attentional layer is followed by dropout with probability $0.5$. For the MLP $\varphi$ with two linear layers, each layer is followed by dropout with probability $0.2$. In all the experiments, we use ADMM as the optimization method. We run $60$ epochs to train our approach.
\color{black}

% There are two ego vehicles in each scenario.
% At each time step, ego vehicles analyze their own sensor data with data from another ego vehicle,
% and identify covisible objects that exist in both views.
% We assume there are no delay in their communication, thus the sensor data is perfectly aligned.

% \begin{figure}
% \centering
%     \includegraphics[width=0.495\textwidth]{figures/fig3.pdf}
%     \caption{The illustration of the CAD simulation. The first row presents 5 urban intersections where we  perform CoID between connected autonomous vehicles. The second row demonstrates examples of RGBD observations obtained by two connected vehicles at the same street intersection from different perspectives.}
%     \label{fig:scenario_demo}
% \end{figure}

For comparison, we first implement a baseline line method that is our full approach but without using GPS information to generate an object location consistency matrix as defined in Eq. (\ref{eq:gi,j}). In addition, we compare our method with  three existing methods as follows:
\begin{itemize}
    \item Graph convolutional neural network for graph matching (\textbf{GCN-GM}) \cite{fey2018splinecnn} that use spline kernel to aggregate objects' themselves and their neighborhood visual-spatial information for graph matching.
    \item Deep graph matching consensus (\textbf{DGMC}) \cite{Fey2020} that performs an iterative refinement process on the similarity matrix given consensus principle.
    \item Bayesian deep graph matching (\textbf{BDGM}) \cite{gao2021bayesian} that performs deep graph matching under Bayesian framework and reduces non-covisible objects based on correspondences uncertainties.
\end{itemize}
None of the comparison methods are capable of integrating GPS information and only ours and BDGM explicitly address non-covisible objects in CoID.

As we treat the CoID as a data retrieval process, we use the following  metrics to evaluate the CoID performance.
\begin{itemize}
    \item \textbf{Precision} is defined as the ratio of the retrieved correct correspondences over all the retrieved correspondences.
    \item \textbf{Recall} is defined as the ratio of the retrieved correct correspondences over the ground truth correspondences.
    \item \textbf{F1 Score} is a metric to evaluate the overall performance of CoID methods, which is defined as $(2\times Precision \times Recall)/(Precision +Recall)$.
\end{itemize}

\subsection{Results over Connected Autonomous Driving Simulations}

The CAD simulation includes a variety of technical challenges to perform CoID, including various street objects (e.g., pedestrians and vehicles) with ambiguous visual appearance, a large number of non-covisible objects, strong occlusion in the perception, noisy observation caused long-distance observing, as well as the realistic noisy GPS information, which follows the specification of the real-world GNSS. We run our approach on a Linux machine with an i7 16-core CPU and 16G memory. The average execution time is around $20$Hz.

\begin{table}[htb]
\centering
\tabcolsep=0.4cm
\caption{Quantitative results of our approach and comparisons with three previous methods over the CAD simulations.}
\label{tab:QuanResults}
\begin{tabular}{|l|c|c|c|}
\hline
Method & Precision & Recall & F1\\
\hline\hline
GCN-GM \cite{fey2018splinecnn} & 0.5001 & 0.6391 & 0.5611 \\
\hline
DGMC \cite{fey2019deep} & 0.4736 & 0.6425 & 0.5453\\
\hline
BDGM \cite{gao2021bayesian}& 0.6817 & 0.6097 & 0.6437\\
\hline\hline
{Ours $w/o$ GPS} & 0.7464  & 0.8006  &  0.7726\\
\hline
{Ours} & \textbf{0.7859} & \textbf{0.8278} & \textbf{0.8063}\\
\hline	
\end{tabular}
\end{table}

\begin{figure*}[t]
\vspace{6pt}
\centering
\subfigure[GCN-GM ]{\includegraphics[width=0.245\textwidth]{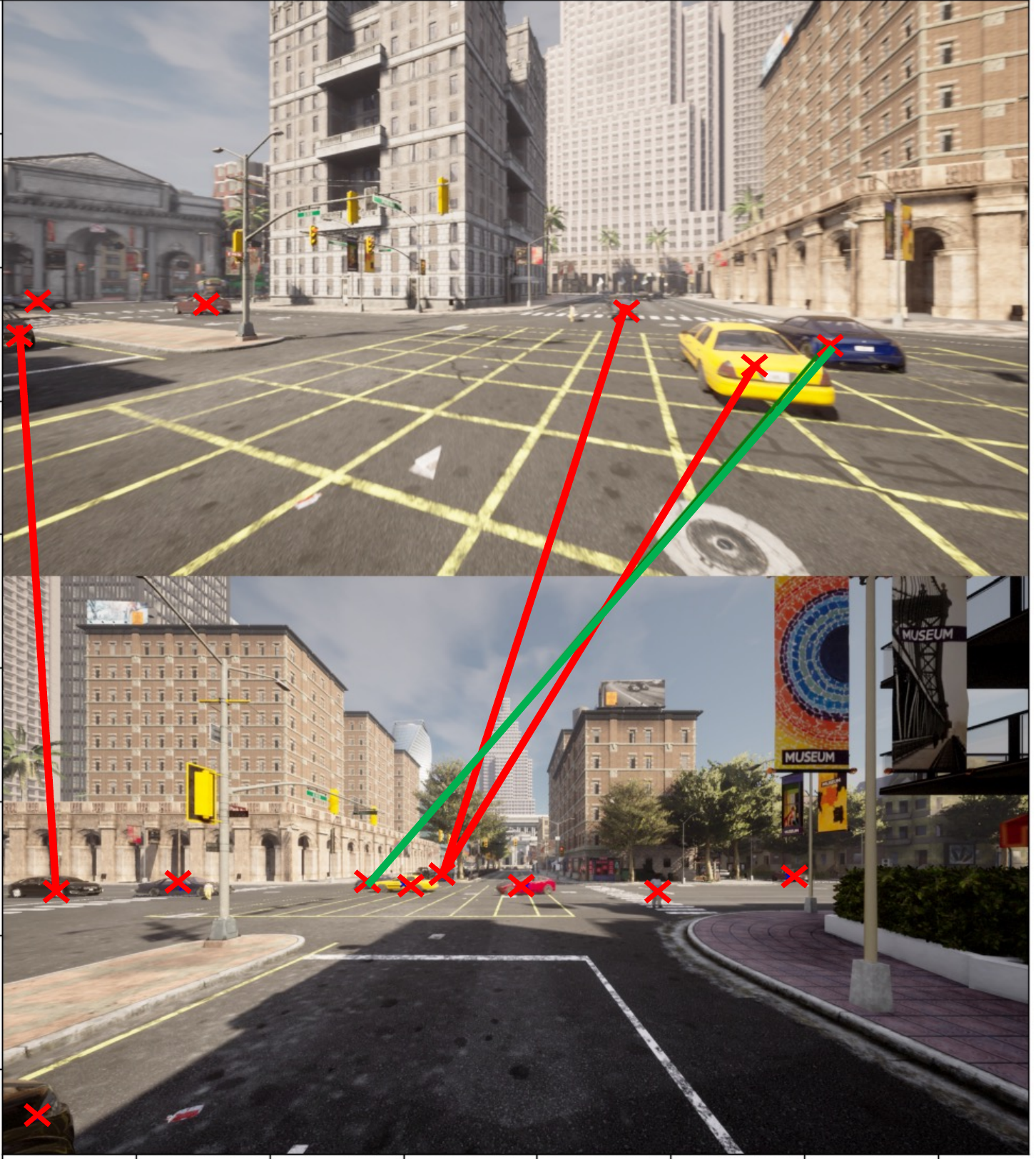}\label{fig:gcn}}
\subfigure[DGMC ]{\includegraphics[width=0.245\textwidth]{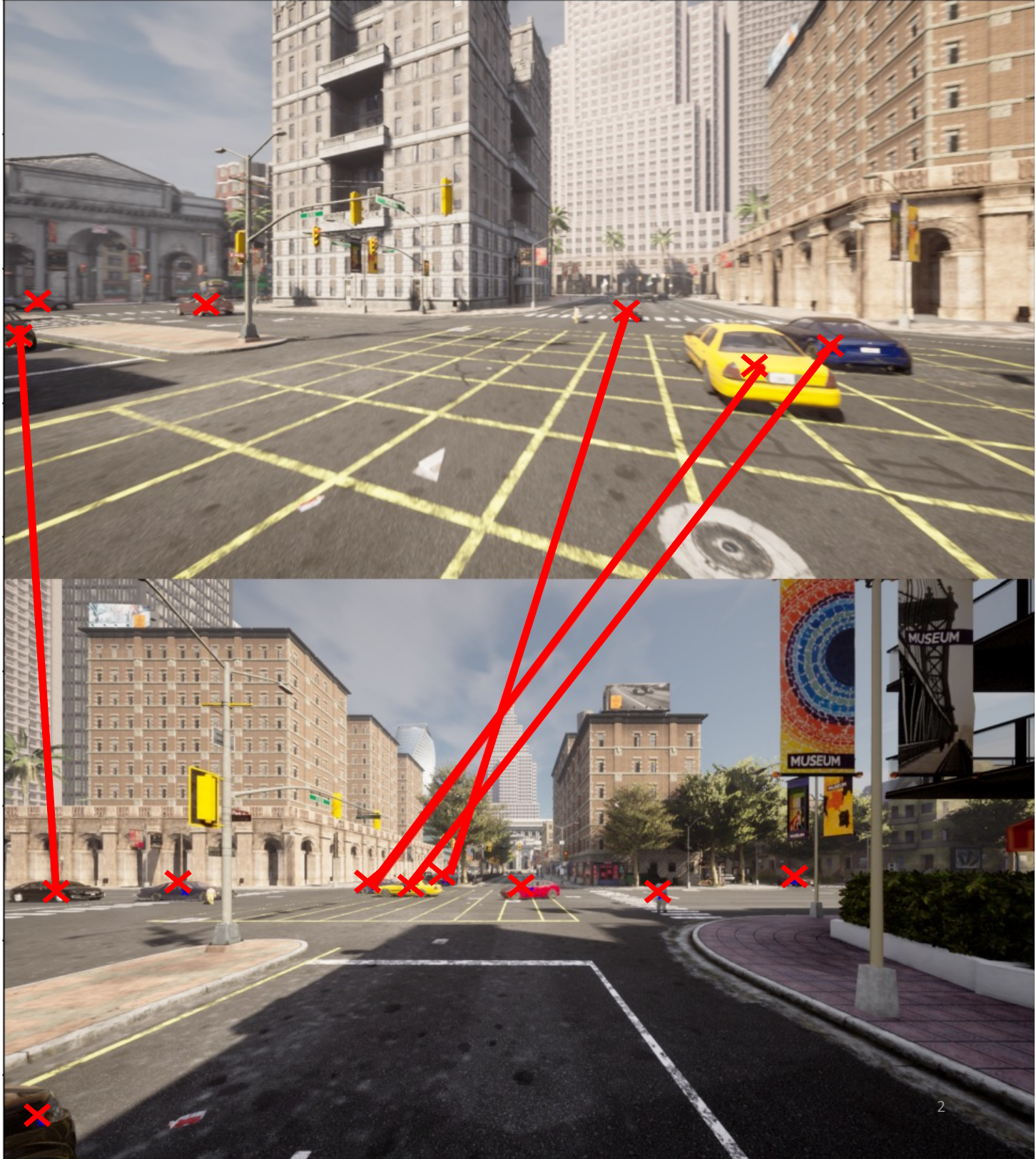}\label{fig:dgmc}}
\subfigure[{BDGM}]{\includegraphics[width=0.245\textwidth]{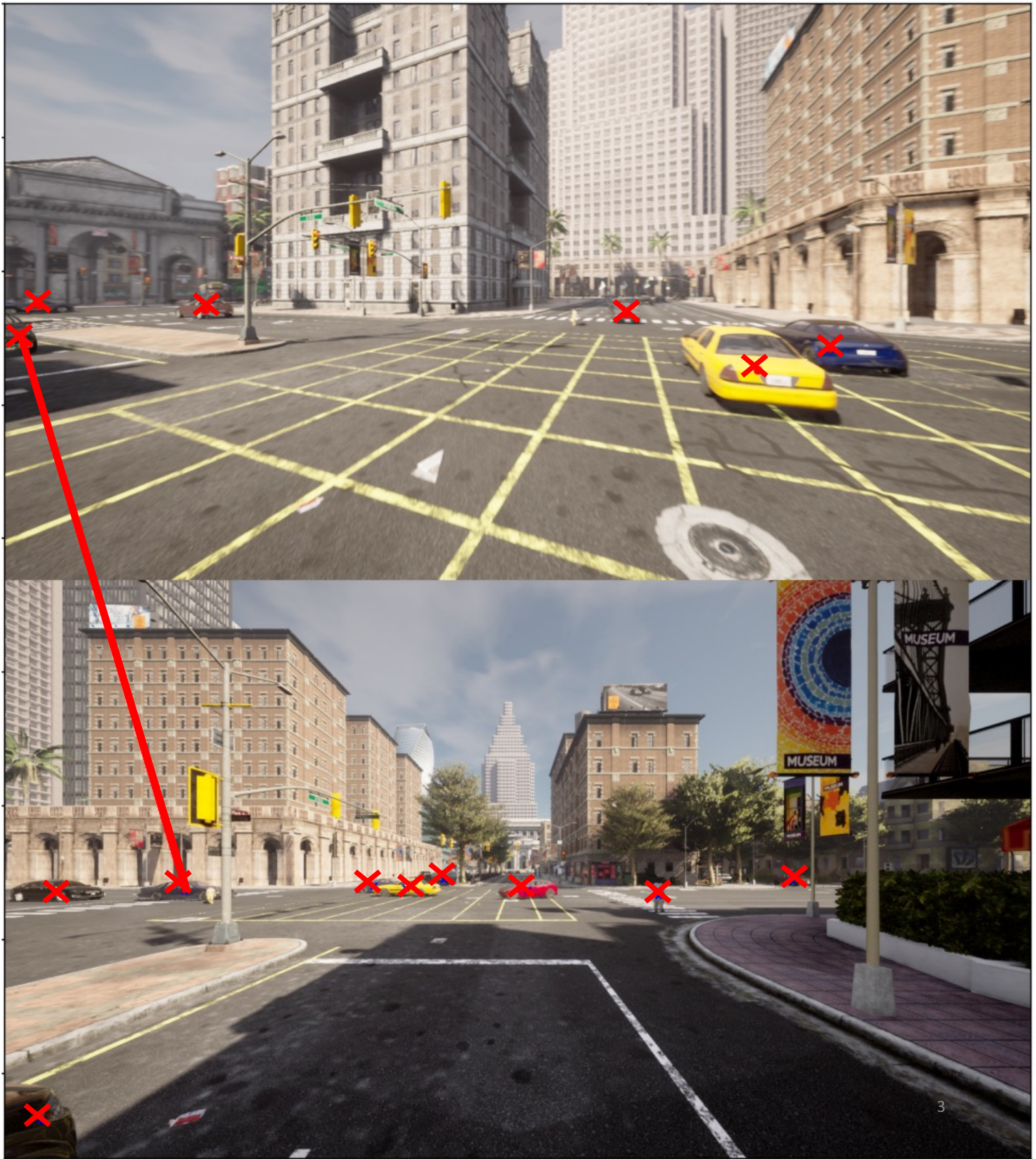}\label{fig:bdgm}}
\subfigure[{Ours}]{\includegraphics[width=0.245\textwidth]{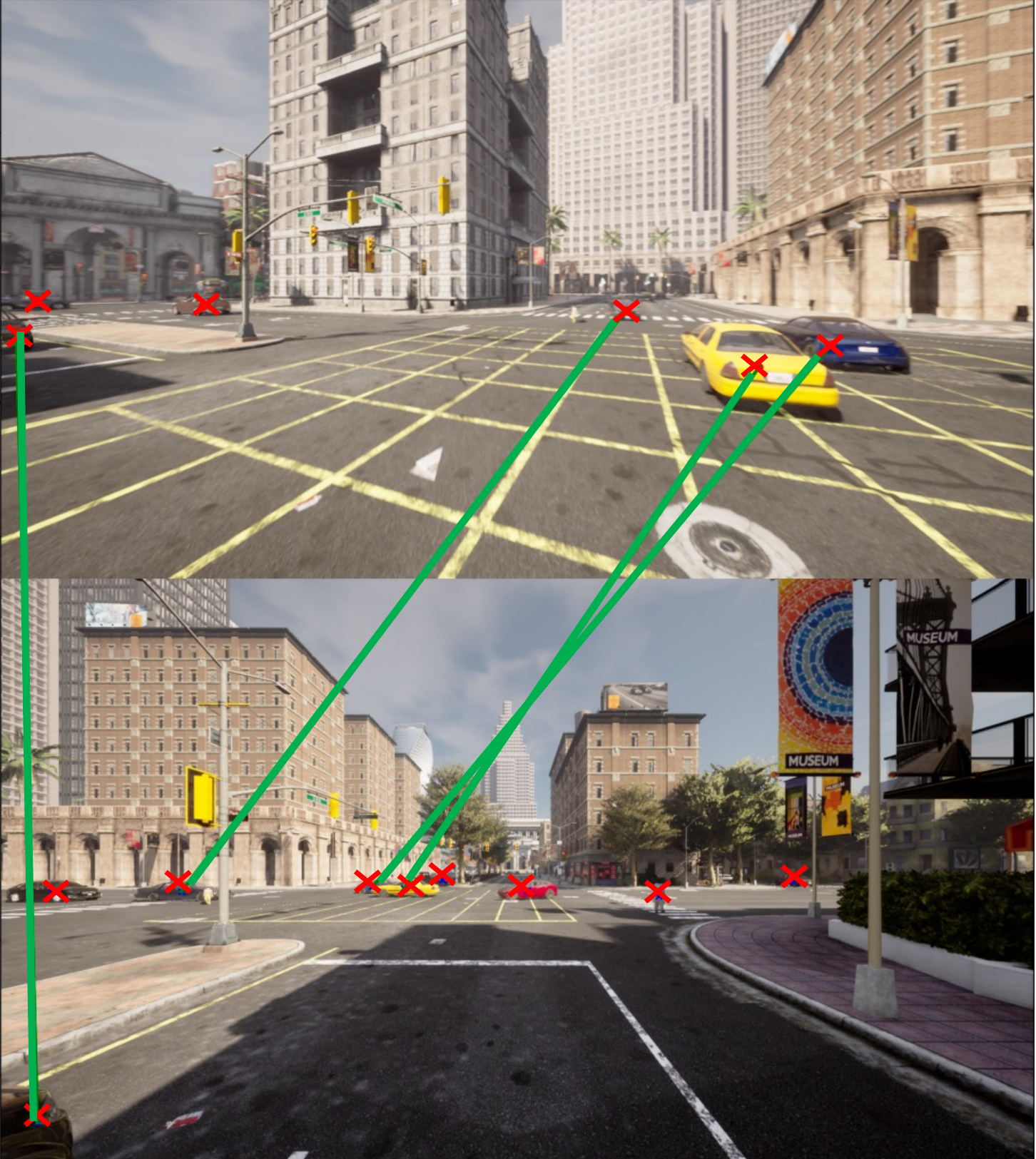}\label{fig:ours}}
\caption{Qualitative experimental results obtained by our approach in the CAD simulations, and comparisons with GCN-GM, DGMC and BDGM. Green lines denote correct correspondences and red lines denote incorrect correspondences. Red cross symbols denote the detected street objects in different views of the vehicles. [Best viewed in color.]}
\label{fig:QualResults}
\end{figure*}

\begin{figure*}[h]
	\centering
\subfigure[Precision-Recall Curve]{\includegraphics[width=0.246\textwidth]{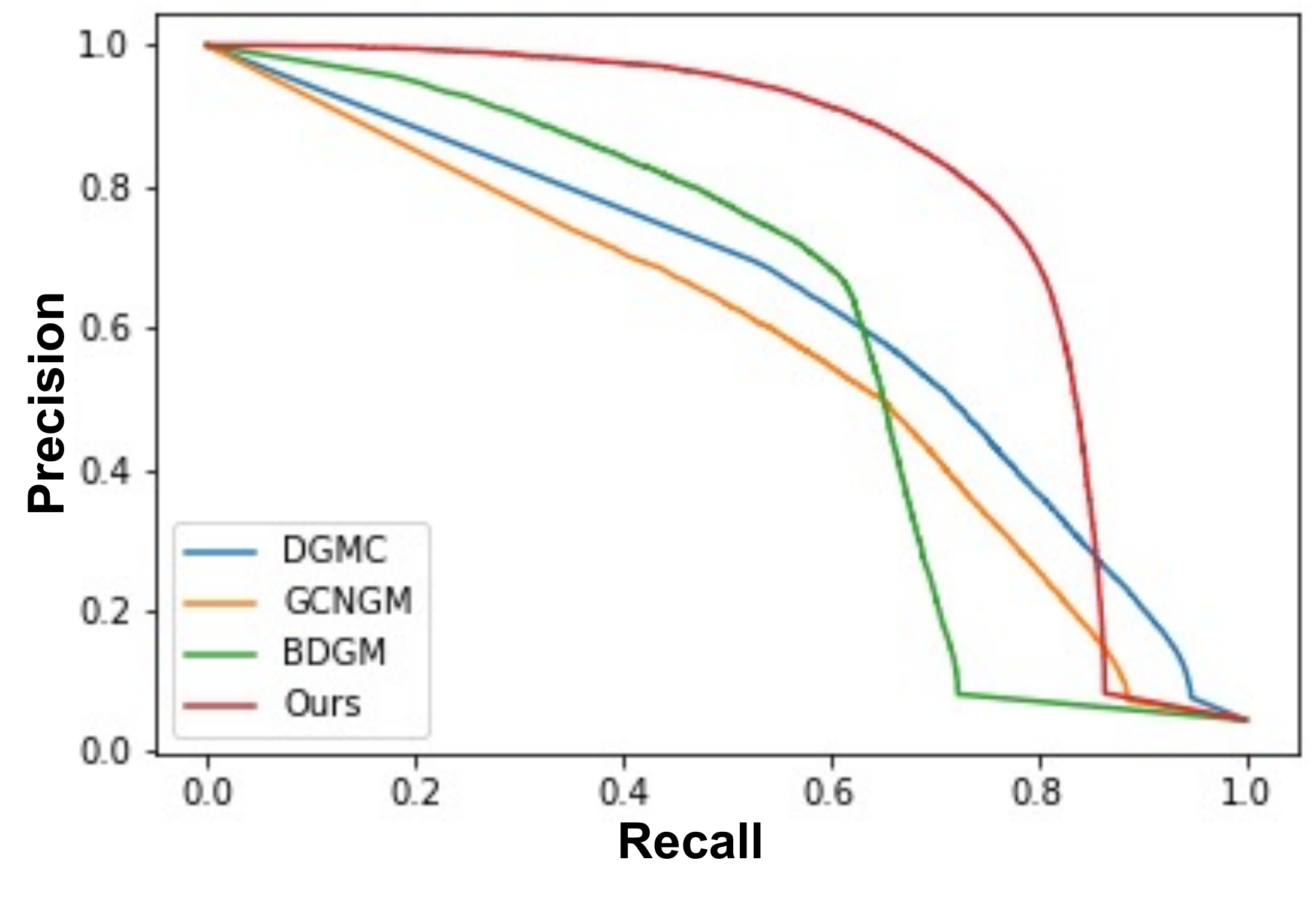}\label{fig:precision-recall}}
\subfigure[Noisy Depth]{\includegraphics[width=0.246\textwidth]{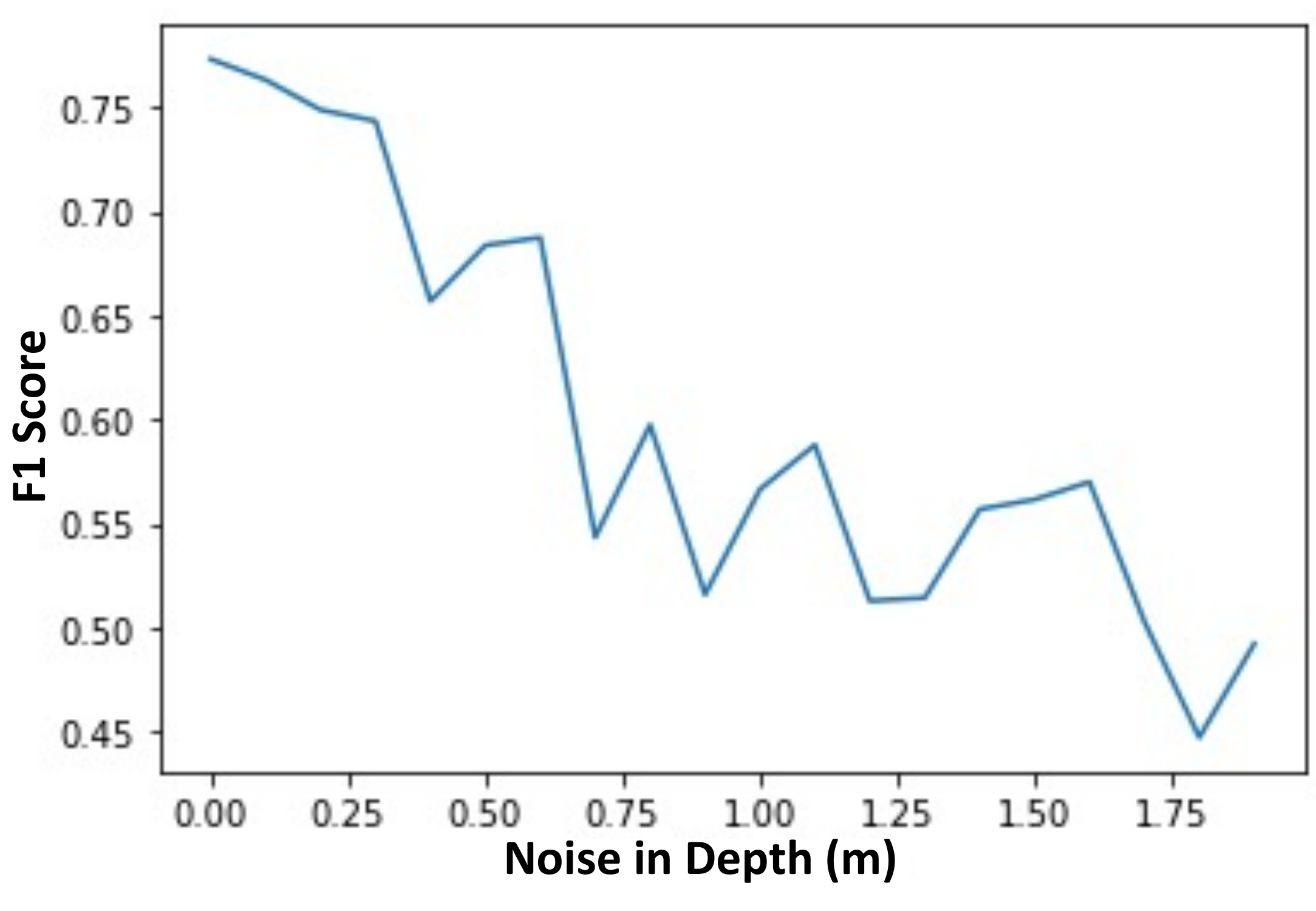}\label{fig:noiseD}}
\subfigure[Noisy GPS]{\includegraphics[width=0.246\textwidth]{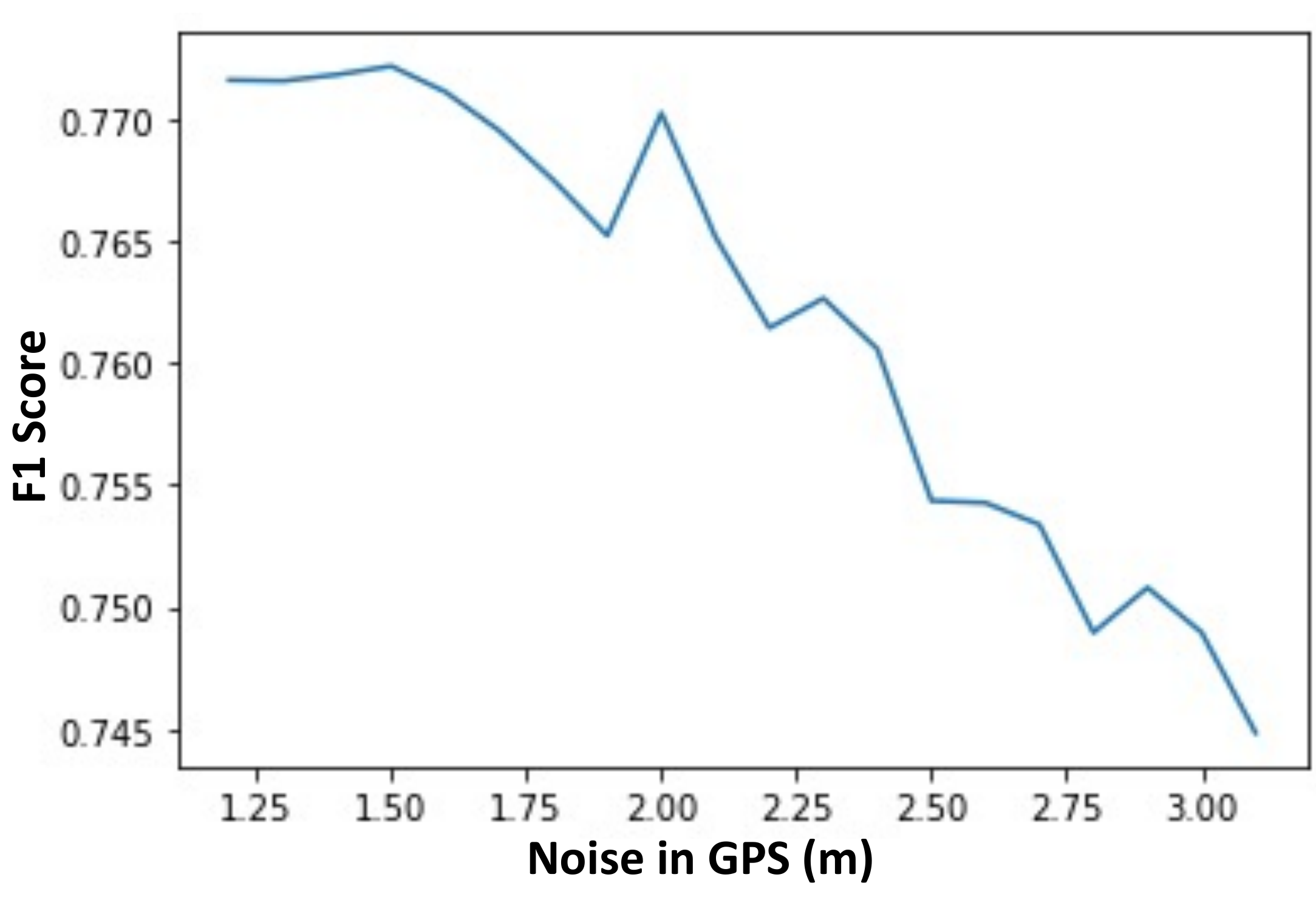}\label{fig:noiseG}}
\subfigure[Varying Thresholds]{\includegraphics[width=0.246\textwidth]{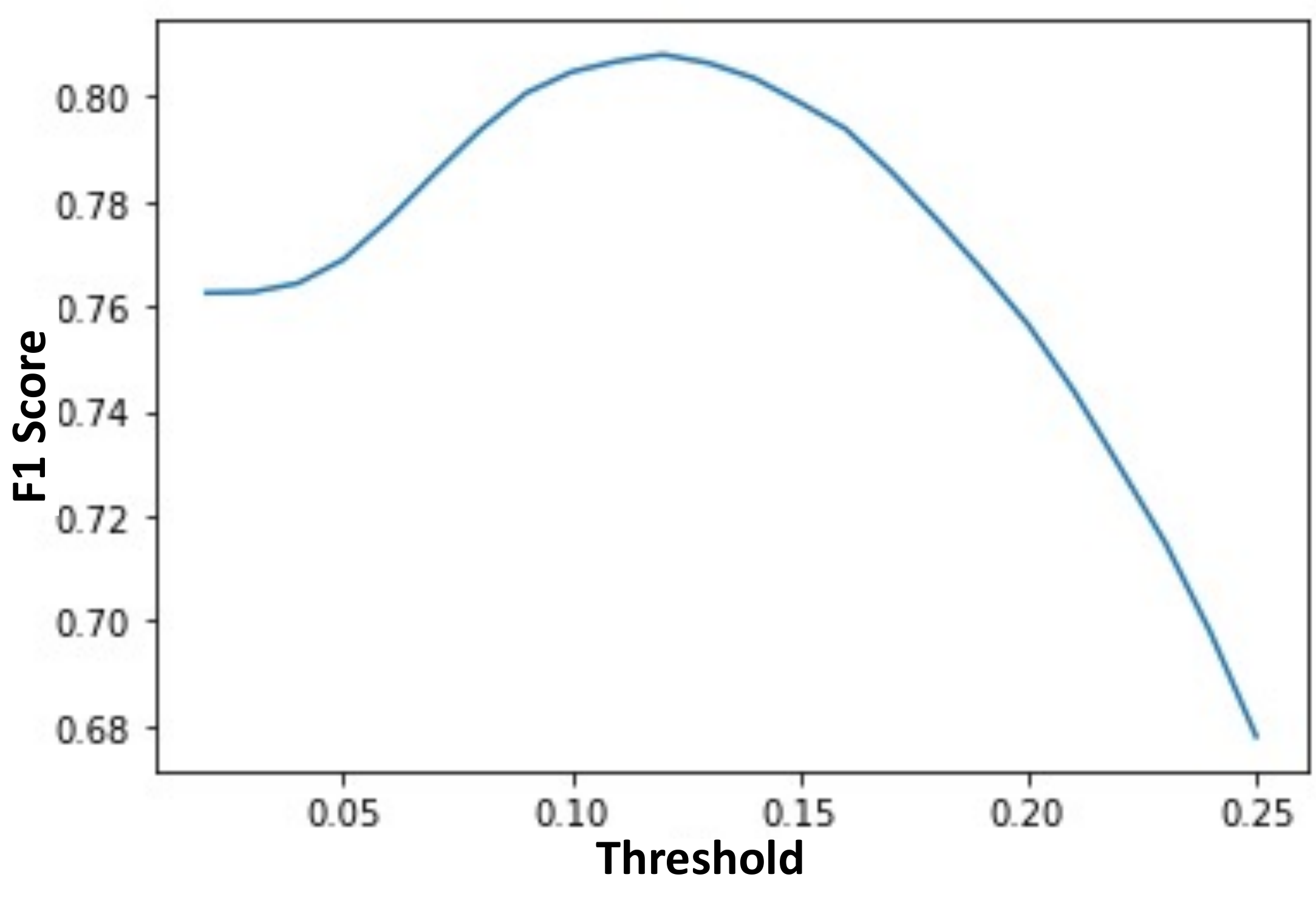}\label{fig:thresh}}
\caption{Discussion of our CoID approach's characteristics, including the precision-recall curve, effects of noise in depth and GPS on the performance, and the analysis of threshold values.
}
\label{fig:discussion}
\end{figure*}

The quantitative results are presented in Table \ref{tab:QuanResults}.
We can see that our baseline approach outperforms all the previous methods on precision, recall, and F1 score due to its capability of addressing non-covisible objects and integrating visual-spatial information for CoID. In addition, our full approach outperforms the baseline method, which indicates the importance of integrating GPS information. In the comparison with previous approaches, GCN-GM performs the worst as it is generally only focusing on integrating visual-spatial features of objects, and does not consider graph pruning and non-covisible object elimination. DGMC performs better than GCN-GM, as it is the first to propose graph matching consensus principle, the final identified correspondences can be refined by updating the consensus. BDGM achieves promising results on precision due to its consideration of removing non-covisible objects given correspondence uncertainties. By efficiently addressing visual ambiguity and non-covisible objects, as well as integrating GPS into graph matching, our approach achieves the best CoID performance.

The qualitative results of our approach are shown in Figure \ref{fig:QualResults}. We can clearly see that our approach correctly identified the correspondences of objects observed by connected vehicles. By comparing with other methods, we can also see that GCN-GM and DGMC perform badly as they aim to maximize the number of correspondences and ignore the influence caused by non-covisible objects. BDGM also performs badly as it removes most of the uncertain correspondences that are caused by sensing noise and low-resolution observations. By integrating multi-modal sensing information and addressing non-covisible objects, our approach can identify correspondences robustly in the CAD scenario.

\subsection{Discussion}
We further study our approach's characteristics, including the overall performance based on the precision-recall curve, robustness to the depth noise and the GPS noise, as well as the analysis of hyper-parameter threshold $\theta$.

\subsubsection{Overall Performance} In order to evaluate the overall performance of CoID, we draw the precision-recall curve as shown in Figure \ref{fig:precision-recall}. We also use a single-value evaluation metric of Area Under the Curve (AUC) to evaluate the overall performance, which 
is defined as the area under the precision-recall curve. Its value is between [0, 1] with a greater value indicating better performance. 
% AUC $=1$ indicates the perfect performance.
It is observed that our approach obtains the AUC of $0.79$, which is significantly larger than the BDGM with $0.6$ and DGMC with $0.64$.

\subsubsection{Robustness to Noise}
Figure \ref{fig:noiseD} demonstrates the effect of varying noise in the sensing depth. We can see that as the increase of depth noise, the performance of our approach gradually decreases with small fluctuation. We can also see that our method obtains robust performance with $0.5m$ depth sensing noise.
Figure \ref{fig:noiseG} illustrates the effect of varying noise in the GPS. We can see that our approach generally shows good robustness to the noise in the GPS information. The GPS noise increasing from $1.2m$ to $3m$ cause $\%2.5$ decrease in the overall performance.
\subsubsection{Threshold Analysis}
We use hyperparameter $\theta$ to threshold the identified correspondences based upon the SoftMax variance in order to remove non-covisible objects. Figure \ref{fig:thresh} shows the sensitivity analysis of the performance influenced by $\theta$ based on the F1 score. We observe that our approach achieves the best performance when $\theta = 0.13$.

\section{Conclusion}\label{sec:conclusion}

Correspondence identification is essential for collaborative perception in connected autonomous driving, with the goal of enabling consistent reference of street objects by connected vehicles.
To address the key technical challenges, including perceptual aliasing, non-covisibility, and noisy perception,
we introduce a novel deep masked graph matching approach for CoID.
Through integrating multi-modal sensing information (including visual, spatial and GPS cues), 
our approach is able to robustly identify the correspondences of street objects.
In addition, we implement a new technique to remove non-covisible objects by thresholding the SoftMax variance.
Finally, we implement a connected autonomous driving simulator by integrating CARLA and SUMO,
and employ it to collect a large-scale dataset that includes around 70K pairs of high-fidelity paired observations with ground truth correspondences for the training and evaluation of CoID methods.
The experimental results show our approach outperforms previous methods and achieves state-of-the-art CoID performance in connected autonomous driving applications.

\bibliographystyle{IEEEtran}
\bibliography{ref}
\end{document}